\documentclass[letterpaper, 10 pt, conference]{ieeeconf}  
\IEEEoverridecommandlockouts
\usepackage{cite}
\usepackage{amsmath,amssymb,amsfonts}
\usepackage{algorithmic}
\usepackage{graphicx}
\usepackage{textcomp}
\usepackage{xcolor}
\usepackage{float}
\usepackage{bm}
\usepackage{mathtools}
\usepackage{multirow}

\usepackage[utf8]{inputenc}  
\usepackage[T1]{fontenc}  
\usepackage{subcaption}
\usepackage{caption}
\DeclareUnicodeCharacter{0308}{HERE!HERE!}

\usepackage{amsmath}
\usepackage{amssymb}
\usepackage{color}
\usepackage{nicefrac}
\usepackage{graphicx}
\usepackage{float}
\usepackage{bm}

\graphicspath{{figures/}}

\title{\LARGE \bf
Contact Sensing via Joint Torque Sensors and a Force/Torque Sensor for Legged Robots}

\author{Jared Grinberg$^{1}$ and Yanran Ding$^{1}$%
\thanks{$^{1}$Jared Grinberg and Yanran Ding are with the Department of Robotics, University of Michigan, Ann Arbor, MI - 48109, USA.
        {\tt\small \{grinberg, yanrand\}@umich.edu}}%
}

\begin{document}

\maketitle
\global\csname @topnum\endcsname 0
\global\csname @botnum\endcsname 0
\thispagestyle{empty}
\pagestyle{empty}

\begin{abstract}
This paper presents a method for detecting and localizing contact along robot legs using distributed joint torque sensors and a single hip-mounted force-torque (FT) sensor using a generalized momentum-based observer framework. We designed a low-cost strain-gauge-based joint torque sensor that can be installed on every joint to provide direct torque measurements, eliminating the need for complex friction models and providing more accurate torque readings than estimation based on motor current. Simulation studies on a floating-based 2-DoF robot leg verified that the proposed framework accurately recovers contact force and location along the thigh and shin links. Through a calibration procedure, our torque sensor achieved an average 96.4\% accuracy relative to ground truth measurements. Building upon the torque sensor, we performed hardware experiments on a 2-DoF manipulator, which showed sub-centimeter contact localization accuracy and force errors below 0.2\,N.
\end{abstract}

\section{Introduction}
Agile legged animals can navigate cluttered environments by detecting contact not just at the foot, but also along the leg. Modern legged robots~\cite{Hutter2017, DiCarlo2018} have demonstrated how contact sensing at the foot can improve performance on uneven terrains. While foot sensors work well for detecting ground contact, additional contact detection on the thigh and shank can further improve locomotion performance regarding obstacle clearance in challenging environments~\cite{Haddadin2017,Iskandar2024}. Failure to detect contact along the robot leg can destabilize the system, especially in cluttered settings where undetected contact induces significant disturbance to the robot.

\begin{figure}[t]
    \centering
    \includegraphics[width=0.9\columnwidth]{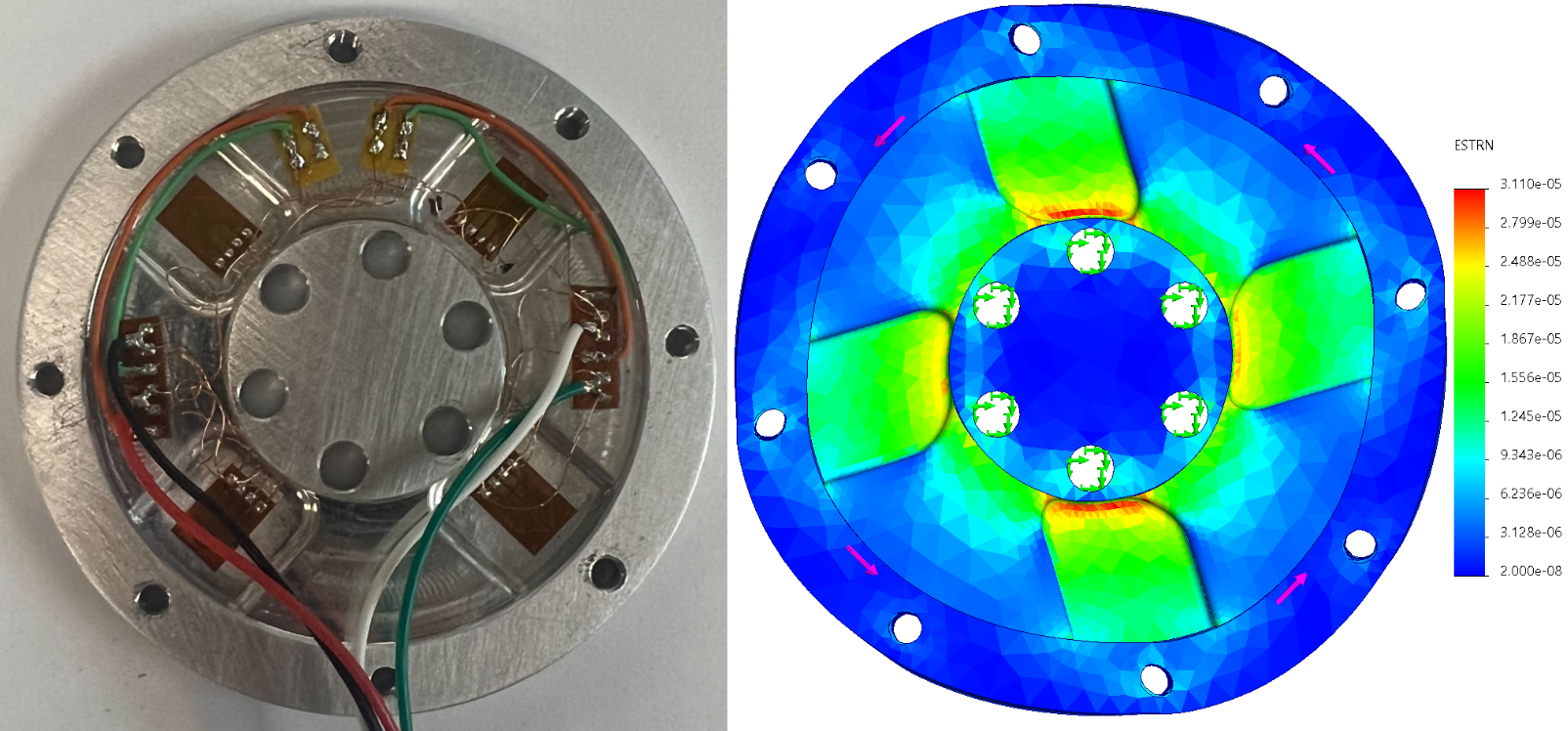}
    \caption{Custom low-cost strain-gauge joint torque sensor. (Left) Fabricated sensor with strain gauges. (Right) Finite-Element Analysis (FEA) of the custom joint torque sensor under 0.4\,Nm load.}
    \label{fig:joint_torque_sensor}
    \vspace{-10pt}
\end{figure}

\begin{figure*}[t]
    \centering
    \includegraphics[width=0.9\textwidth]{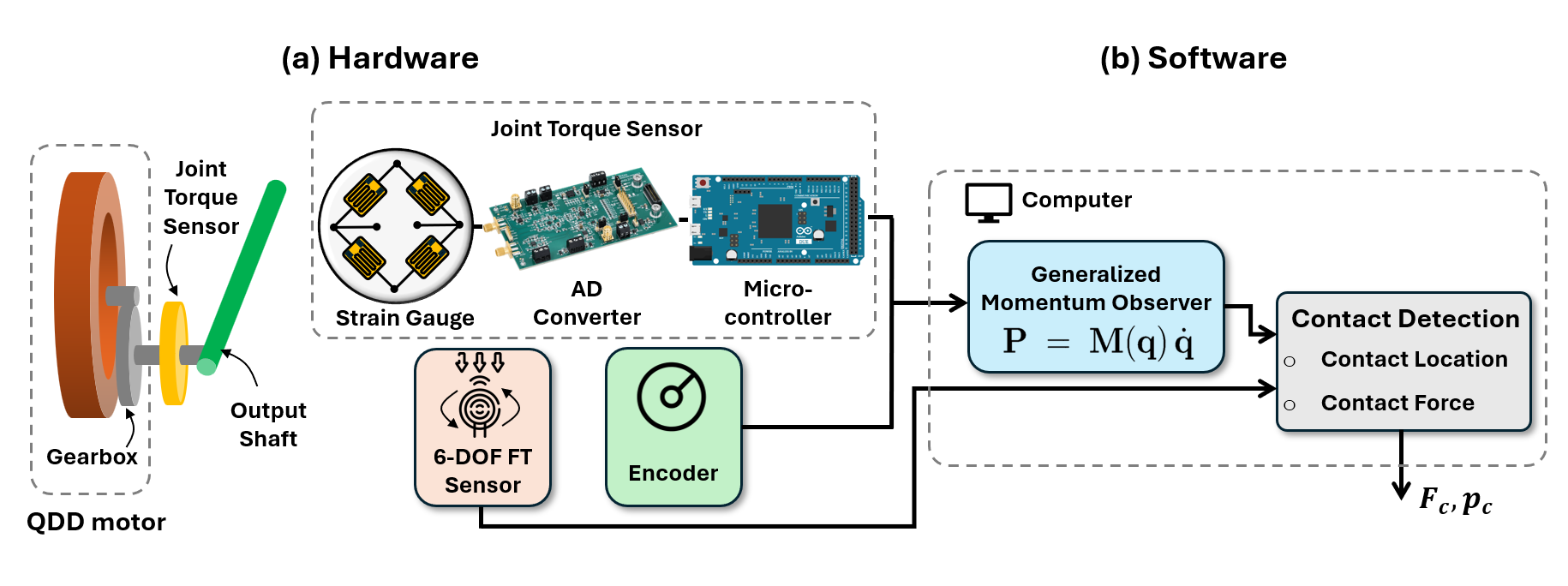}
    \caption{System schematic depicting data flow for contact detection: (a) motor hardware with joint torque sensors, data acquisition system, and encoders, which feed into (b) a generalized momentum observer sensing algorithm, subsequently input with a 6-DoF force-torque (FT) sensor data for contact detection, outputting estimated contact force and location.}
    \label{fig:flow_diagram}
    \vspace{-10pt}
\end{figure*}

Many contact detection algorithms use motor current measurements to indirectly infer contact. \cite{Haddadin2017,Zurlo2023,Le2013} use current sensing to estimate joint torque; \cite{Haddadin2008,Wahrburg2018} use observer-based methods dependent on motor current; \cite{Wahrburg2015} incorporate the friction model into a momentum-based Kalman filter, where friction and gearbox nonlinearities adversely affect torque estimation~\cite{Armstrong-Helouvry1994}, especially in the high torque region. Generalized momentum-based observers~\cite{DeLuca2005,Haddadin2017} can detect collisions reliably, yet localizing contact on a particular link is difficult with unmodeled friction. \cite{Bledt2018} uses 
contact model fusion to improve reliability in detecting foot contact, but does not sensorize the rest of the leg. The contact particle filter~\cite{Manuelli2016,Zurlo2023} has contact localization capability along the robot surface, at the cost of increased computation. 

Other contact detection strategies use alternative sensing modalities such as tactile skins~\cite{Cheng2019}, whether it be vision-based~\cite{Cheng2019, Yuan2017} or barometer-based~\cite{Hou2025}. While tactile sensors offer rich information and large coverage across robot limbs, these solutions often have a slower update rate. Scaling tactile sensors for multi‐link collision detection requires the use of extensive sensing arrays, which results in fragile hardware that is not suitable for high-impact conditions common in legged robots.

Contact detection and localization algorithms that rely on motor current sensing require accurate estimation of the joint output torque. While estimating joint torque from motor current is good enough for quasi-direct-drive (QDD) motors~\cite{seok2014design} due to its low gear ratio, actuators with moderately high gear ratio often face issues including nonlinear friction, backlash, hysteresis and torque ripple that contaminate torque estimation. Measuring torque directly at the joint output bypasses drivetrain friction entirely~\cite{Taghirad2000,Peng2021}, offering high-bandwidth and precise readings for collision detection. Nevertheless, challenges exist before actuators equipped with torque sensors are widely adopted to legged robots, such as the ability to endure high impact, while satisfying stringent design specifications. 


On top of the hardware challenges, detecting and localizing contact with joint torque sensors alone presents difficulties. That is because the contact information accumulates further down the kinematic chain from the base, while upstream links lack sufficient joint torque readings to fully determine the contact. A force-torque (FT) sensor may be mounted at the base to provide sufficient sensor data for proximal links. Although FT sensor may be fragile under high impact, the intermediate links between the foot and sensor act as a suspension system for protection. 

\subsection{Contributions}
Our work presents a framework for detecting and localizing contacts along a robot leg by leveraging direct joint torque sensors and a hip-mounted FT sensor. Our contributions are three-fold:
\begin{enumerate}
    \item A low-cost custom strain-gauge-based torque sensor (shown in Fig. \ref{fig:joint_torque_sensor}) that demonstrates high accuracy torque sensing;
    \item Implementation of a generalized momentum-based observer algorithm that fuses joint torque sensor and FT sensor data to detect and localize collision along the robot leg;
    \item Hardware experimental validation on a 2-link leg that demonstrates high sensing accuracy.
\end{enumerate}

The remainder of this paper is organized as follows: Section~\ref{sec:methods} presents the momentum-based observer framework, Section~\ref{sec:contact_localization} details the contact localization method, Sections~\ref{sec:simulationstudy} and~\ref{sec:hardwareexperiment} present simulation and hardware validation, followed by discussion and conclusion.

\vspace{.5em}
\section{Momentum-Based Observer}
\label{sec:methods}

This section introduces the robot dynamics equations of motion and the generalized momentum-based observer, which is the backbone of the contact detection algorithm. This section corresponds to the Generalized Momentum Observer block in Fig. \ref{fig:flow_diagram}(b).

\subsection{Robot Dynamics}
The Equation of Motion (EoM) for fixed-base robots:
\begin{equation}
    \mathbf{M}(\mathbf{q}) \,\ddot{\mathbf{q}}
    \;+\;
    \mathbf{C}\bigl(\mathbf{q}, \dot{\mathbf{q}}\bigr)\,\dot{\mathbf{q}}
    \;+\;
    \mathbf{g}\bigl(\mathbf{q}\bigr)
    \;+\;
    \boldsymbol{\tau}_\mathrm{ext}
    \;=\;
    \boldsymbol{\tau}_\mathrm{sen},
    \label{eq:dynamics_no_friction}
\end{equation}
where $\mathbf{q}\in\mathbb{R}^{n}$ is the generalized joint coordinates; $\dot{\mathbf{q}}, \ddot{\mathbf{q}} \in\mathbb{R}^{n}$ are generalized velocity and acceleration; $\mathbf{M}\in\mathbb{R}^{n\times n}$ is the inertia matrix; $\mathbf{C}\dot{\mathbf{q}}\in\mathbb{R}^{n}$ is the Coriolis/centrifugal term; $\mathbf{g}\in\mathbb{R}^{n}$ is the gravity vector; $\boldsymbol{\tau}_\mathrm{ext}\in\mathbb{R}^{n}$ is the external torque/force mapped into joint space; $\boldsymbol{\tau}_\mathrm{sen}\in\mathbb{R}^{n}$ is the measured joint torque. Note that $\boldsymbol{\tau}_\mathrm{sen}$ can only be directly sensed if there is a joint torque sensor at the output shaft.

Equation \eqref{eq:dynamics_no_friction} is challenging to use directly for collision detection in practice because estimating $\ddot{\mathbf{q}}$ from noisy velocity data often requires finite difference or filtering. Furthermore, inverting $\mathbf{M}(\mathbf{q})$ can be computationally intensive for robots with high degrees-of-freedom (DoF). These challenges motivate the adoption of generalized momentum for contact detection.


\subsection{Generalized Momentum}
The use of generalized momentum $\mathbf{P} = \mathbf{M}(\mathbf{q})\,\dot{\mathbf{q}}$ bypasses the explicit use of $\ddot{\mathbf{q}}$ and allows computation based on measurable torques and known system properties. Taking the time derivative and substituting from equation \eqref{eq:dynamics_no_friction} gives:
\begin{align}
    \dot{\mathbf{P}}
    \;=\;&
    \frac{d}{dt}\bigl[\mathbf{M}(\mathbf{q})\,\dot{\mathbf{q}}\bigr]
    \;=\;
    \mathbf{M}(\mathbf{q})\,\ddot{\mathbf{q}}
    \;+\;
    \dot{\mathbf{M}}(\mathbf{q})\,\dot{\mathbf{q}}
    \nonumber \\[4pt]
    =\;&
    \boldsymbol{\tau}_\mathrm{sen}
    \;-\;
    \boldsymbol{\tau}_\mathrm{ext}
    \;-\;
    \mathbf{C}\,\dot{\mathbf{q}}
    \;-\;
    \mathbf{g}
    \;+\;
    \dot{\mathbf{M}}\;\dot{\mathbf{q}}.
    \label{eq:Pdot_full}
\end{align}

Using the property \(\dot{\mathbf{M}} = \mathbf{C} + \mathbf{C}^\top\) \cite{Haddadin2017}, we get:
\begin{equation}
    \dot{\mathbf{P}}
    \;=\;
    \mathbf{C}^\top \dot{\mathbf{q}}
    \;-\;
    \mathbf{g}
    \;+\;
    \boldsymbol{\tau}_\mathrm{sen}
    \;-\;
    \boldsymbol{\tau}_\mathrm{ext}.
    \label{eq:Pdot_raw}
\end{equation}

This reformation isolates the only unknown external torque term while the rest of the terms are measurable. Let us define
$\mathbf{u} \;\;=\;\; \mathbf{C}^\top \dot{\mathbf{q}} \;-\; \mathbf{g} \;+\; \boldsymbol{\tau}_\mathrm{sen},$
we get
$\dot{\mathbf{P}}
\;=\;
\mathbf{u}
\;-\;
\boldsymbol{\tau}_\mathrm{ext}
\label{eq:Pdot_u}$, where the term $\mathbf{u}$ represents the predicted momentum rate of change based on system dynamics and measured actuator torques. When no external contact occurs, $\mathbf{u}$ should equal $\dot{\mathbf{P}}$, making any deviation from it a direct indicator of external wrench. Collision detection is achieved by monitoring the residual:
\begin{equation}
    \mathbf{r}(t)
    \;=\;
    \mathbf{K} \,\Bigl[\,
        \mathbf{P}(t)
        \;-\;
        \mathbf{p}_\mathrm{int}(t)
        \;-\;
        \mathbf{P}_0
    \Bigr],
    \label{eq:residual}
\end{equation}
where $\mathbf{K} \in \mathbb{R}^{n \times n}$ is a diagonal gain matrix; $\mathbf{p}_\mathrm{int} \in \mathbb{R}^{n}$ is the observer’s internal momentum estimate (integrator state); and $\mathbf{P}_0 \in \mathbb{R}^{n}$ is the initial momentum value \cite{DeLuca2005}. The residual $\mathbf{r}(t)$ serves as the collision detection signal by comparing the actual momentum $\mathbf{P}(t)$ against the predicted momentum from the observer $\mathbf{p}_\mathrm{int}(t)$. When external forces are absent, these should match and $\mathbf{r}(t) \approx 0$. Discretizing in time yields:
\begin{align}
    \mathbf{p}_\mathrm{int}(t + \Delta t)
    \;=\;
    \mathbf{p}_\mathrm{int}(t)
    \;+\;
    \Bigl[\mathbf{u}
      \;+\;
      \mathbf{r}(t)
    \Bigr]\Delta t.
    \label{eq:pint_update}
\end{align}

When no external torque is present, \(\mathbf{r}(t)\approx 0\). Collision is declared when $\|\mathbf{r}(t)\|$ exceeds a threshold $\epsilon_{\text{res}}$.

\subsection{Collision Link Identification}

The link on which contact occurs is identified by observing the residual. When a collision takes place on link $c$, the residual vector follows:
\begin{subequations} \label{eq:link_r}
\begin{align}
    & r_i(t)\,\neq\,0 
    \;\; \text{for}\; i=1,\dots,c; 
    \\
    & r_j(t)\,=\,0
    \;\; \text{for}\; j=c+1,\dots,n.
\end{align}
\end{subequations}

To determine which link has been impacted, we observe that for an open kinematic chain of $n$ serial links, a collision on link $c$ causes the first $c$ residual components $r_1, r_2,\dots,r_c$, causing them to rise above zero, while the remaining residuals, $r_{c+1},\dots,r_{n}$, corresponding to more distal links, remain (approximately) zero throughout the duration of the contact. This follows from the fact that only the joints from $1$ to $c$ lie in the force transmission path between the contact point and the base, while distal joints ($c+1$ to $n$) do not carry that load and thus corresponding residuals remain unaffected \cite{Haddadin2017,DeLuca2005}. We identify the contacted link by finding the highest indexed residual that exceeds the detection threshold:
\begin{equation}
    c 
    \;=\;
    \max\{\, i \in \{1,\dots,n\} : |r_i(t)| > \epsilon_{\text{res}} \}.
    \label{eq:link_c}
\end{equation}

\begin{figure}[t]
    \centering
    \vspace{5pt}
    \includegraphics[width=0.45\columnwidth]{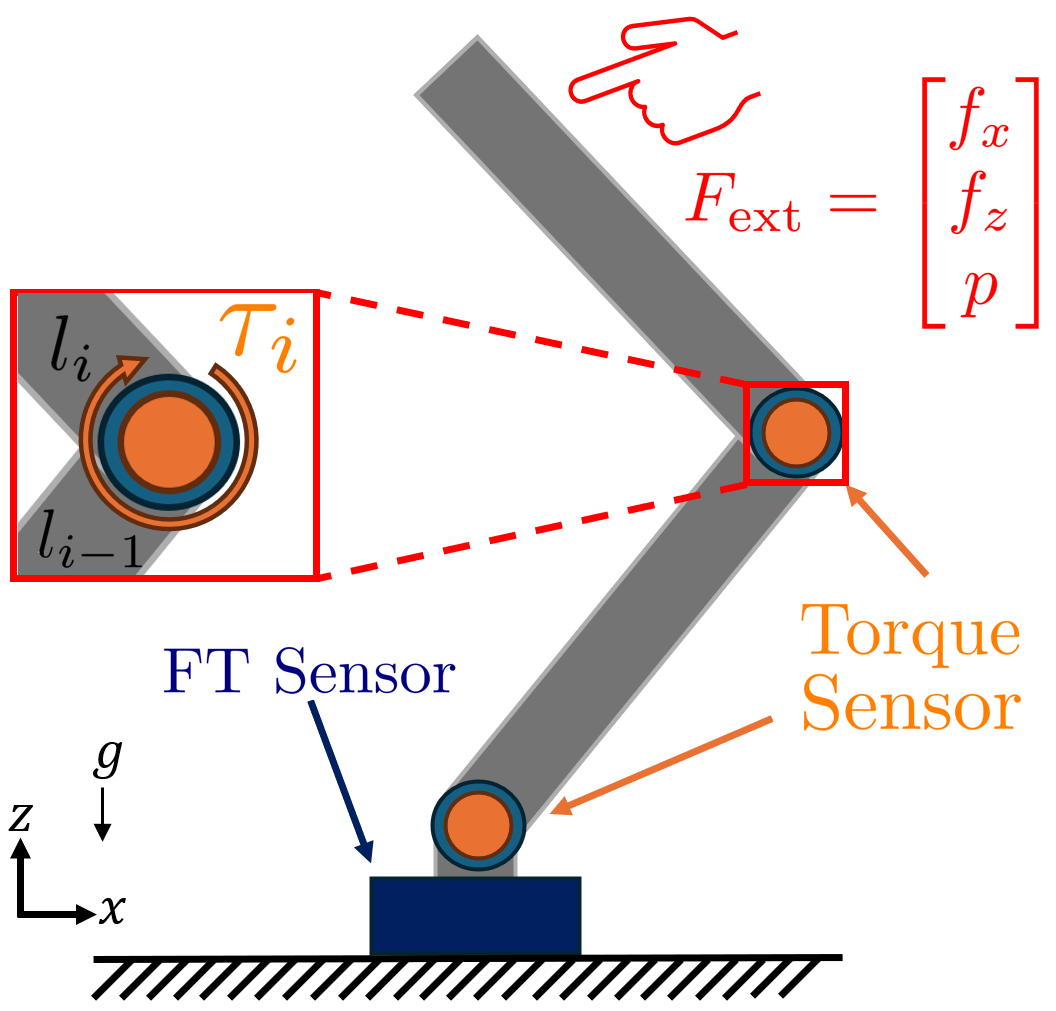}
    \caption{Sensor configuration rationale for multi-link collision detection. A planar point contact has 3 degrees of freedom (Fx, Fz, and contact position $p$), but a 2-DoF serial chain provides only 2 joint torque measurements ($\tau_1$, $\tau_2$), creating an underdetermined system. The base-mounted FT sensor provides the necessary additional measurements for complete contact localization while remaining effective for contacts on any link.}
    \label{fig:sensors_config_arm}
    \vspace{-10pt}
\end{figure}

\section{Contact Detection}
\label{sec:contact_localization}

The contact detection problem is to identify the force vector $\bm{F}_c$ and the contact location $\bm{p}_c$. This section discusses how to estimate external contact by combining a base FT sensor and distributed joint torque sensors. This section corresponds to the Contact Detection block in Fig. \ref{fig:flow_diagram}(b).

\subsection{Base Force/Torque Sensor}
The effectiveness of joint torque sensors alone for contact sensing diminishes as we go towards the base of the kinematic chain, that is because collision on link $c$ will only be perceived by the preceding links $i=1,\cdots, c$. To solve this issue, we mount a force/torque (FT) sensor at the base to provide additional measurements for proximal links. For example, Fig. \ref{fig:sensors_config_arm} shows a serial 2-DoF manipulator equipped with torque sensors on joint 1 and 2. The external force is characterized by three variables: the force vector $\mathbf{F}_c:=[F_c^x, F_c^z]^\top$ and contact location $p_c$ along the link. Relying on the joint torque reading $[\tau_1, \tau_2]^\top$ is not sufficient to solve for the external force and location. However, the base FT sensor provides additional readings $\mathbf{F}^b:=[F_x^{b},F_y^b,M_y^b]^\top$ for complete contact calculation. The rest of the section will continue using the 2D example for illustrative purposes.

We simulate the base FT sensor by extending the generalized coordinate to include 3 virtual joints $[x,y,z]$, where $x$ and $z$ are prismatic joints, $y$ is a revolute joint perpendicular to the $x-z$ plane, where the generalized joint is $\bm{q}=[x, y, z, q_1, q_2]^\top\in\mathbb{R}^N$. The generalized force applied on the virtual joints while taking into account of the manipulator dynamics is:
\begin{equation}
    \mathbf{w}=\mathbf{S} (\mathbf{C}\dot{\mathbf{q}}+\mathbf{g}-\mathbf{B}\,\boldsymbol{\tau}_\mathrm{sen}),
    \label{eq:Rterm}
\end{equation}
where $\mathbf{w}\in\mathbb{R}^3$ is the generalized force at the FT sensor; $\mathbf{H}\ddot{\mathbf{q}}$ is left out since we assume joint acceleration is negligible; $\mathbf{B} \in \mathbb{R}^{N \times 2}$ maps the actuated joint torques to the generalized force space; the selection matrix $\mathbf{S}\in \mathbb{R}^{3 \times N}=[\mathbf{I}^{3\times 3}, \mathbf{0}^{3\times 2}]$ maps the generalized force to the base FT sensor wrench space. The net unexpected force $\mathbf{F}^u\in\mathbb{R}^3$ is:
\begin{equation}
    \mathbf{F}^u = \mathbf{w} - \mathbf{F}^b
    \label{eq:F_unexp}
\end{equation}
Therefore, the contact force vector $\mathbf{F}_c=-[F_x^u, F_z^u]^\top$.

To determine the contact location $\mathbf{p}_c$, we assume it lies between $\mathbf{p}_1,\mathbf{p}_2 \in \mathbb{R}^2$, which are the start and endpoints of the $c$-th link in world coordinates.
\begin{equation}
    \mathbf{p}_c
    \;=\;
    \mathbf{p}_1
    \;+\;
    \alpha\,(\mathbf{p}_2 - \mathbf{p}_1),
    \quad
    \alpha \in [0,1].
    \label{eq:contact_position}
\end{equation}
Enforcing static moment balance about the base point yields the equation:
\begin{equation}
    M_y^u + \mathbf{p}_c \wedge \mathbf{F}^u_{xz} = 0,
    \label{eq:zero_moment}
\end{equation}
where the operator $\wedge$ is the wedge product. By plugging the expression of $\mathbf{p}_c$ to equation \ref{eq:zero_moment} we can solve for $\alpha$ as
\begin{equation}
\alpha = -\frac{M_y^u + \mathbf{p}_1 \wedge \mathbf{F}^u_{xz}}{(\mathbf{p}_2-\mathbf{p}_1)\wedge \mathbf{F}^u_{xz}}.
    \label{eq:get_alpha}
\end{equation}
If $0 \leq \alpha \leq 1$, the collision lies physically on link $c$.

A commercial six-axis force-torque sensor (PixONE by Bota Systems) is mounted at the hip to measure the net base wrench. We only use its planar components ($F_x, F_z, M_y$) to measure the net external force and moment on the entire 2-DoF limb.

\subsection{Joint Torque Sensor}
One contribution is a low-cost strain-gauge-based torque sensor, as shown in Fig. \ref{fig:joint_torque_sensor}. We use a full Wheatstone bridge configuration of 1\,k$\Omega$ gauges with 5\,V excitation and differential output. The sensor bodies were made of 6061 aluminum. We model the strain-gauge sensitivity as:
\begin{equation}
    \frac{\Delta R}{R} \;=\; \mathrm{GF}\,\epsilon, 
    \quad
    V_o \;=\; V_{\mathrm{ex}} 
             \times 
             \frac{\Delta R / R}{4}
             \quad 
    \text{(quarter-bridge)},
\end{equation}
where $\mathrm{GF}=2$ is the gauge factor, $\epsilon$ is the strain, and $V_{\mathrm{ex}}=5\,\mathrm{V}$ is the excitation. The sensor's bridge sensitivity is 
\(
    S = V_o / \epsilon.
\)
For data acquisition, we employ a 24-bit analog-to-digital converter (ADC) with a reference voltage of $V_{\mathrm{ref}}=2.5\,\mathrm{V}$. Although it has a nominal resolution of $\mathrm{LSB} = V_{\mathrm{ref}} / 2^{24}$, practical noise considerations yield an effective number of bits (ENOB) around 16. Consequently, the theoretical minimal detectable strain becomes
\begin{equation}
    \epsilon_{\min} 
    \;=\; 
    \frac{4 \,V_{\mathrm{ref}}}{\mathrm{GF}\,V_{\mathrm{ex}} \,2^N}
    \;\;\approx\; 
    5.96\times10^{-8} 
    \quad (\text{ideal}),
\end{equation}
but rises to $\sim 1.53\times10^{-5}$ when accounting for ENOB. Our finite-element analysis (FEA) of the aluminum disc shows that expected strains under typical torques (ideal = 0.001 to ENOB = 0.4\,Nm) comfortably fall within this detectable range. In practice, the strain gauges produce a differential amplitude of $\sim$10\,mV. A calibration procedure verified the linearity of the sensors’ responses as:
\begin{align*}
    y &= 0.0115\,x + 5.0069, && \text{with } R^2 = 0.9991,\\
    y &= -0.0108\,x - 2.3260, && \text{with } R^2 = 0.9999,
\end{align*}
over the operational load range (0.0196\,Nm to 1.962\,Nm), confirming minimal hysteresis and good repeatability. Fig.~\ref{fig:joint_torque_sensor} shows the Finite-Element Analysis of the joint torque sensor under 0.4\,Nm of torque along with the final fabricated sensor.


\subsection{Contact Estimation Procedure}
The proposed contact estimation approach leverages friction-agnostic torque sensing $\boldsymbol{\tau}_\mathrm{sen}$ at each joint and a base FT sensor. In real-time operation, we first compute the residual $\mathbf{r}$ from Equation~\eqref{eq:residual} and compare it to a threshold. Once a collision is detected, we identify which link $c$ is contacted using Equation~\eqref{eq:link_c}. Next, we compute the contact force and solve for contact location using Equation~\eqref{eq:F_unexp}.

\section{Simulation Study}
\label{sec:simulationstudy}
The proposed framework is tested in both fixed-based (Sec.~\ref{sec:simulationstudy:simulationfixed}) and floating-base (Sec.~\ref{sec:simulationstudy:simulationfloating}) simulations. We evaluated two distinct collision scenarios:
\begin{itemize}
    \item \textbf{Scenario~1:} A 5~N force was applied on Link 1 at a location $\alpha=0.5$, oriented at an angle of $-\,\pi/3$ rad. 
    \item \textbf{Scenario~2:} A 7~N force was applied on Link 2 at $\alpha=0.8$, oriented at an angle of $-\,\pi/3$ rad.
\end{itemize}
Both simulations are performed in MATLAB using \texttt{ode45}.

The physical parameters of the robot are summarized in Table~\ref{tab:physical}: link lengths ($l$), distances from joints to centers of mass ($r$), link masses ($m$), moments of inertia ($I$), base mass ($m_{\mathrm{base}}$), and the friction coefficient ($\mu$) used in the contact model. Simulation and control parameters are summarized in Table~\ref{tab:control}: virtual spring ($K_b$) and damping ($D_b$) constants, PD control gains ($K_P$, $K_D$), standard deviations of the simulated force-torque sensor noise ($\sigma_{FT}$), and the momentum observer residual threshold ($\epsilon_{res}$) for contact detection, which was determined empirically from hardware tests.

\begin{table}[htb]
\caption{Physical constants of the robot}
\label{tab:physical}
\centering
\begin{tabular}{ccc}
\hline
\textbf{Constant} & \textbf{Value} & \textbf{Unit} \\ 
\hline
$l$ & [0.205, 0.215] & m \\
$r$ & [0.171, 0.031] & m \\
$m$ & [0.351, 0.080] & kg \\
$I$ & [0.00207, 0.00030] & kg$\cdot$m$^2$ \\
$m_{\mathrm{base}}$ & 0.738 & kg \\
$\mu$ & 0.3 & \textbackslash \\ 
\hline
\end{tabular}
\end{table}

\begin{table}[htb]
\vspace{5pt}
\caption{Simulation and control parameters}
\label{tab:control}
\centering
\begin{tabular}{ccc}
\hline
\textbf{Parameter} & \textbf{Value} & \textbf{Unit} \\ 
\hline
$K_b$ & [5000, 5000, 500] & N/m \\
$D_b$ & [50, 50, 20] & Ns/m \\
$K_P$ & 500 & \textbackslash \\
$K_D$ & 10 & \textbackslash \\
$\sigma_{FT}$ & [0.1, 0.1, 0.01] & [N, N, Nm] \\
$\epsilon_{res}$ & 0.06 & \textbackslash \\ 
\hline
\end{tabular}
\vspace{-5pt}
\end{table}


\subsection{Fixed-Base Simulation}
\label{sec:simulationstudy:simulationfixed}

The setup of the fixed-base robot is shown in Fig. \ref{fig:sensors_config_arm}. To emulate a FT sensor, virtual spring-damper are included in the base $(x, z, \theta)$ directions, yielding $\mathbf{F}^b$ that corresponds to the net wrench at the base joint. The joint torque readings $\boldsymbol{\tau}_\mathrm{sen}$ are assumed to be ideal, where complications such as drivetrain friction is negligible. A proportional-derivative (PD) controller running at 1~kHz is used to command the hip and knee to move between the two joint configurations, while collisions are introduced at specified times, magnitudes, and link locations.

Table~\ref{tab:simResultsfixed} summarizes the mean and standard deviation of the errors in force and contact position during the contact phase. In both tests, the observer detects the collision after $t=0.5\,\mathrm{s}$ and localizes the contact point to within a few millimeters on a leg measuring roughly 200\,mm per link. Force errors below 0.15\,N.

\begin{table}[htb]
\centering
\caption{Results of Fixed-Base Simulation}
\label{tab:simResultsfixed}
\resizebox{\columnwidth}{!}{%
\begin{tabular}{l|ccc|ccc}
\hline
\multirow{2}{*}{\textbf{Scenario}} & \multicolumn{3}{c|}{\textbf{Force Errors (N)}} & \multicolumn{3}{c}{\textbf{Position Errors (mm)}} \\
 & $\text{Fx}$ & $\text{Fz}$ & $|\mathbf{F}|$ & $p_x$ & $p_z$ & $|\mathbf{p}|$ \\ 
\hline
\textbf{Test 1} Mean & -0.008 & -0.008 & 0.112 & 1 & 0 & 3 \\
\hspace{22pt} STD & 0.089 & 0.090 & 0.058 & 4 & 1 & 2 \\
\hline
\textbf{Test 2} Mean & 0.002 & -0.006 & 0.142 & 0 & 2 & 2 \\
\hspace{22pt} STD & 0.108 & 0.111 & 0.059 & 2 & 1 & 1 \\
\hline
\end{tabular}%
}
\end{table}

\vspace{-10pt}



\begin{figure}[H]
    \centering
    \includegraphics[width=0.47\columnwidth]{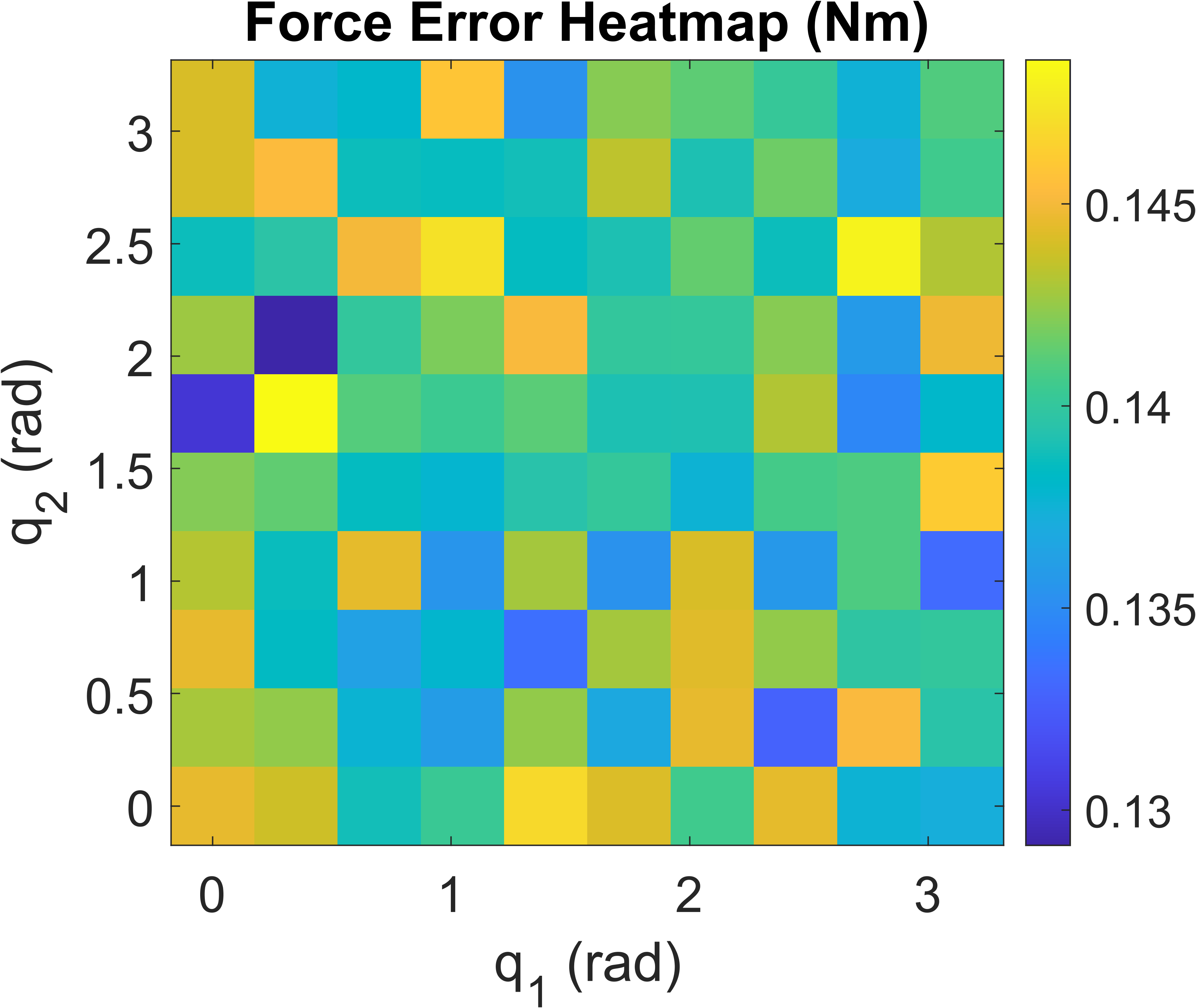}
    \includegraphics[width=0.45\columnwidth]{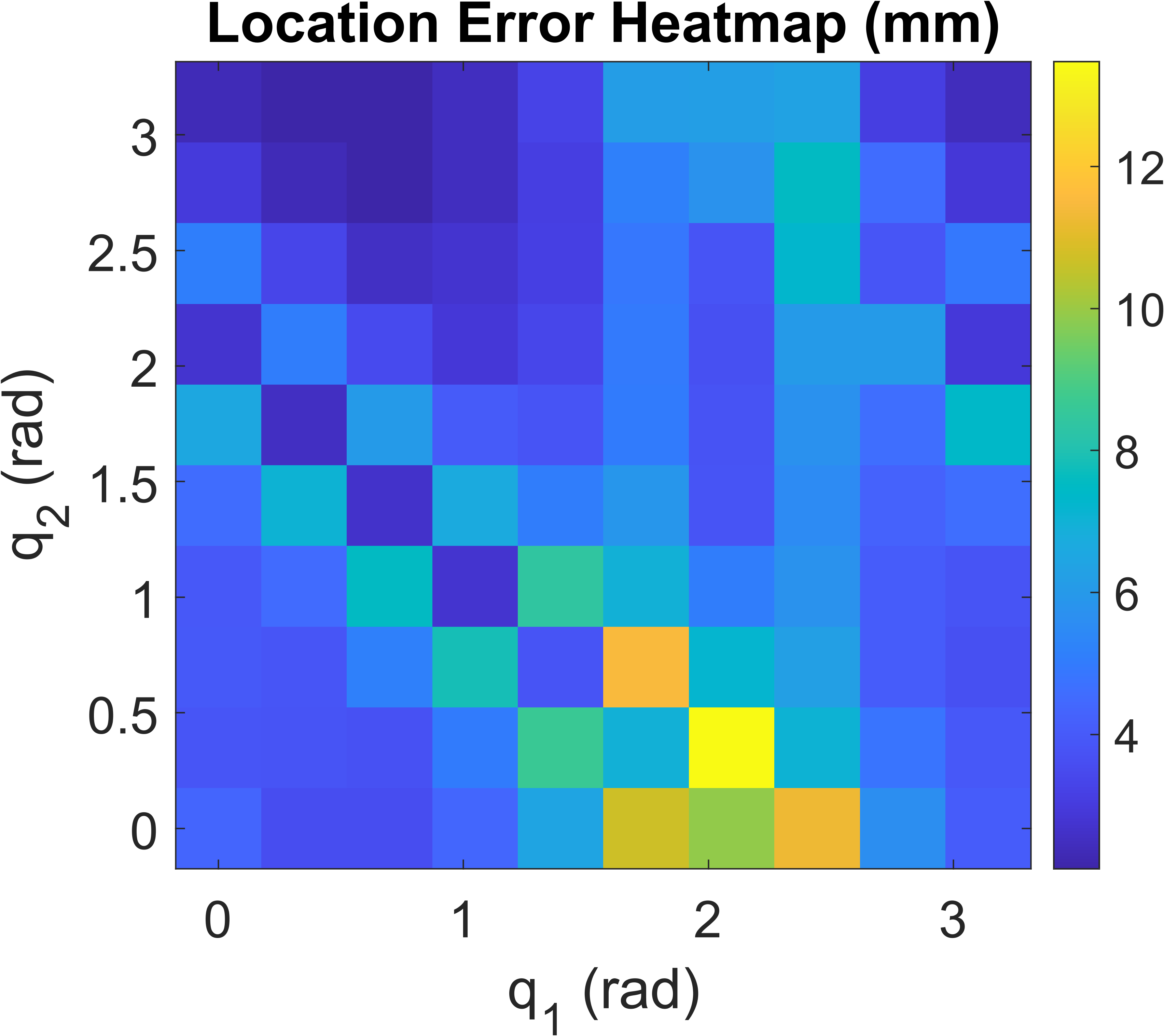}
    \caption{Results of Fixed-Base Simulation Parametric Sweep of Configurations.}
    \label{fig:simResultsfixed_100_sims}
    \vspace{-10pt}
\end{figure}


To evaluate how the robot configuration influences collision-estimation accuracy, we perform a parametric sweep of joint angles $\{q_1,q_2\}$ over a $10\times10$ grid in the range $[0,\pi]\times[0,\pi]$. For each $(q_1,q_2)$ pair, we simulate contacts placed at four different points along each link ($\alpha \in \{0.25, 0.5, 0.75, 1.0\}$) using a known force vector. Across all 100 configurations, the location error remains below $13.5\,\mathrm{mm}$ (approximately $7\%$ of the link length), and the force error below $0.15\,\mathrm{N}$ (about $2\%$ of the applied force). These results, shown in Fig.~\ref{fig:simResultsfixed_100_sims} indicate that our observer-based approach consistently maintains sub-centimeter precision regardless of the specific leg posture. Notably, the location error heatmap exhibits a visible line of slightly elevated error values across a subset of configurations. This artifact arises when the applied external force becomes nearly collinear with the local link geometry, leading to an ambiguous solution for the contact point. In such rare cases, the solution becomes more sensitive to small perturbations in force or configuration, leading to reduced estimation accuracy.


\begin{figure}[t]
\vspace{5pt}
    \begin{subfigure}{\columnwidth}
    \vspace{5pt}
        \centering
        \includegraphics[width=0.4\columnwidth]{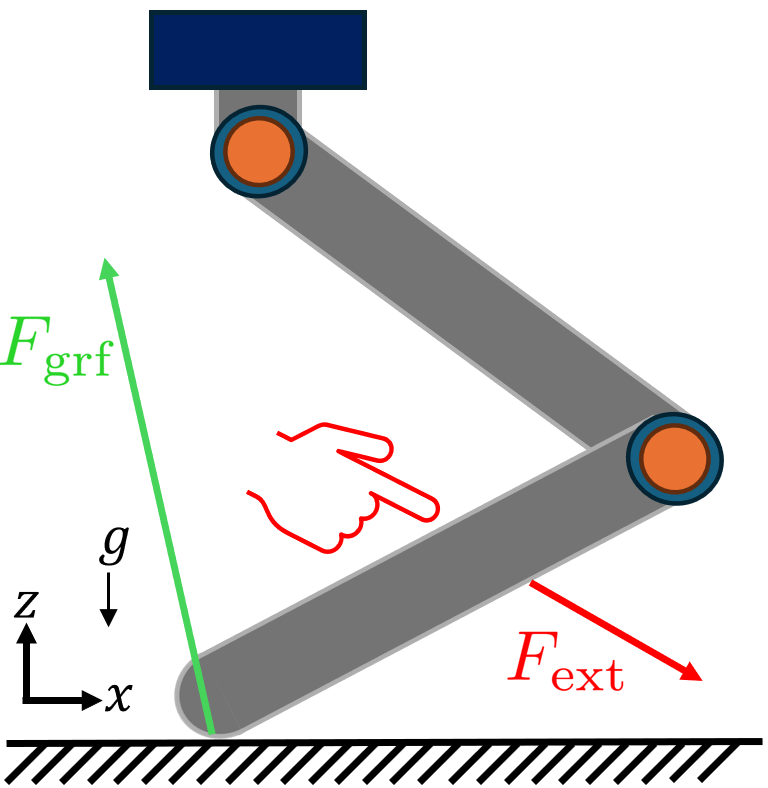}
        \caption{Simulation setup for the floating-base 2-DoF leg.}
        \label{fig:simsetup}
    \end{subfigure}
    \vspace{10pt}

    \begin{subfigure}{\columnwidth}
        \centering
        \includegraphics[width=1\columnwidth]{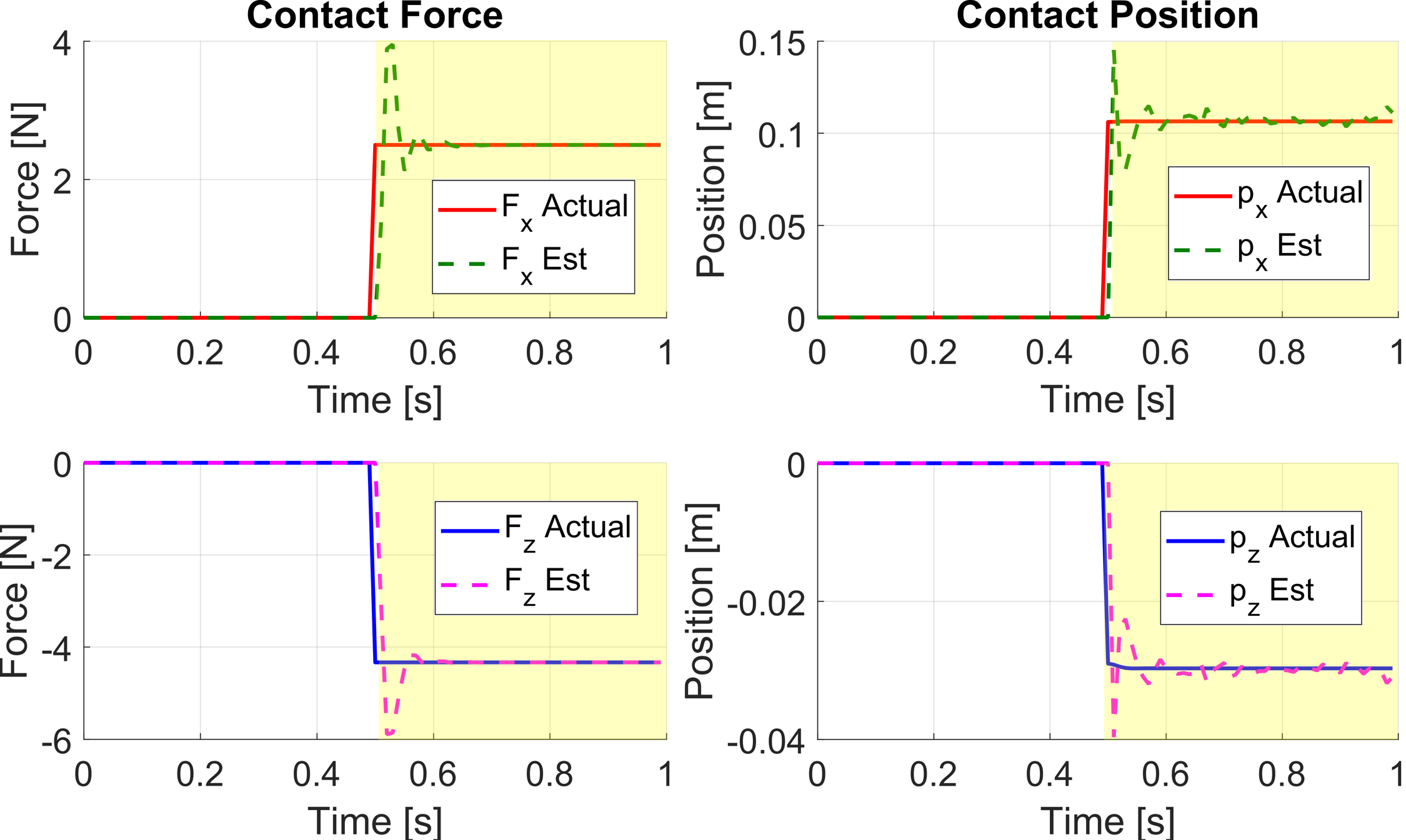}
        \caption{Floating Base, Scenario 1: Collision on Link 1 ($5\,\mathrm{N},\,\alpha=0.5$)}
        \label{fig:scen1subfloating}
    \end{subfigure}
    
    \vspace{10pt}
    
    \begin{subfigure}{\columnwidth}
        \centering
        \includegraphics[width=1\columnwidth]{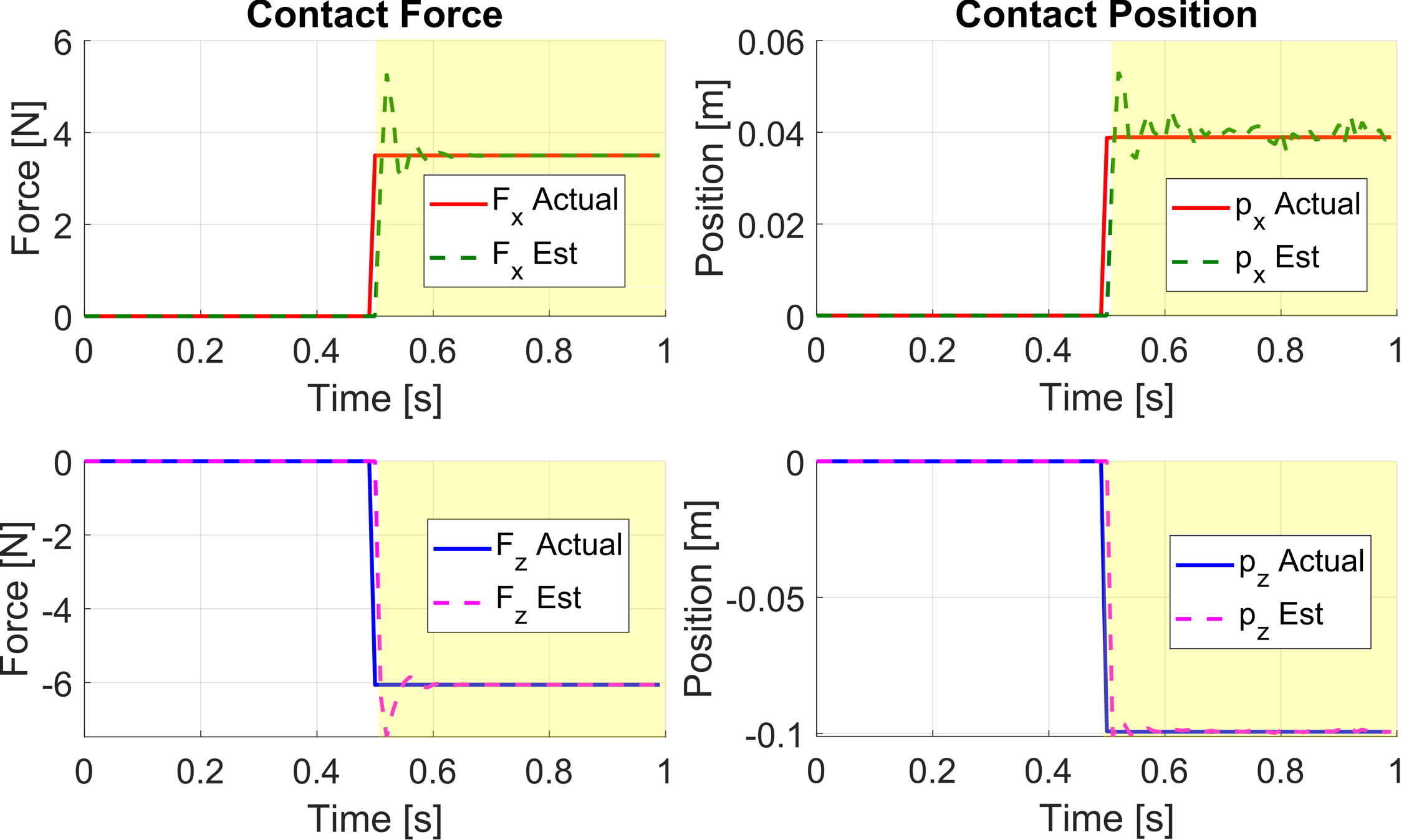}
        \caption{Floating Base, Scenario 2: Collision on Link 2 ($7\,\mathrm{N},\,\alpha=0.8$)}
        \label{fig:scen2subfloating}
    \end{subfigure}

    
    
    

    \caption{Collision detection for a floating-base robot. 
    (a) Illustration of the setup with red external force and green ground reaction force.
    (b) and (c) Show actual vs. estimated contact force magnitude and location, yellow regions indicate post-collision response (t $\geq$ 0.5s).}
    \label{fig:allsimulationresults}
    \vspace{-10pt}
\end{figure}

\subsection{Floating-Base Simulation}
\label{sec:simulationstudy:simulationfloating}

The proposed framework is validated on a floating-base legged robot. The planar robot has a torso that can move in the $x$ and $z$ axes, with its pitch fixed. The robot stands on the ground with a joint PD controller while external forces are applied to its links. Fig.~\ref{fig:simsetup} presents the experiment setup, where the external force (red arrow) is identified aside from the ground reaction force (green arrow). Fig.~\ref{fig:scen1subfloating} and \ref{fig:scen2subfloating} show the collision detection performance, with the yellow highlighted regions showing post-collision response, demonstrating that the observer can track both contact force and location throughout the collision event.

\begin{table}[htb]
\centering
\caption{Results of Floating-Base Simulation}
\label{tab:simResultsfloating}
\resizebox{\columnwidth}{!}{%
\begin{tabular}{l|ccc|ccc}
\hline
\multirow{2}{*}{\textbf{Scenario}} & \multicolumn{3}{c|}{\textbf{Force Errors (N)}} & \multicolumn{3}{c}{\textbf{Position Errors (mm)}} \\
 & $\text{Fx}$ & $\text{Fz}$ & $|\mathbf{F}|$ & $p_x$ & $p_z$ & $|\mathbf{p}|$ \\ 
\hline
\textbf{Test 1} Mean & -0.033 & 0.051 & 0.162 & -1 & 0 & 5 \\
\hspace{22pt} STD & 0.340 & 0.379 & 0.486 & 8 & 2 & 7 \\
\hline
\textbf{Test 2} Mean & -0.036 & 0.050 & 0.121 & -1 & 0 & 2 \\
\hspace{22pt} STD & 0.313 & 0.236 & 0.378 & 4 & 1 & 3 \\
\hline
\end{tabular}%
}
\vspace{-10pt}
\end{table}

Table~\ref{tab:simResultsfloating} summarizes the error metric during the collision phase for the floating-base simulation. The average force estimation error remains below 0.17\,N, with a higher variance compared to the fixed-base results. The increased variance is likely caused by transient dynamics as the system stabilizes after contact in the floating-base setup.



\section{Hardware Experiments}
\label{sec:hardwareexperiment}

\subsection{Hardware Setup}
\begin{figure}[H]
    \centering
    \includegraphics[width=0.8\columnwidth]{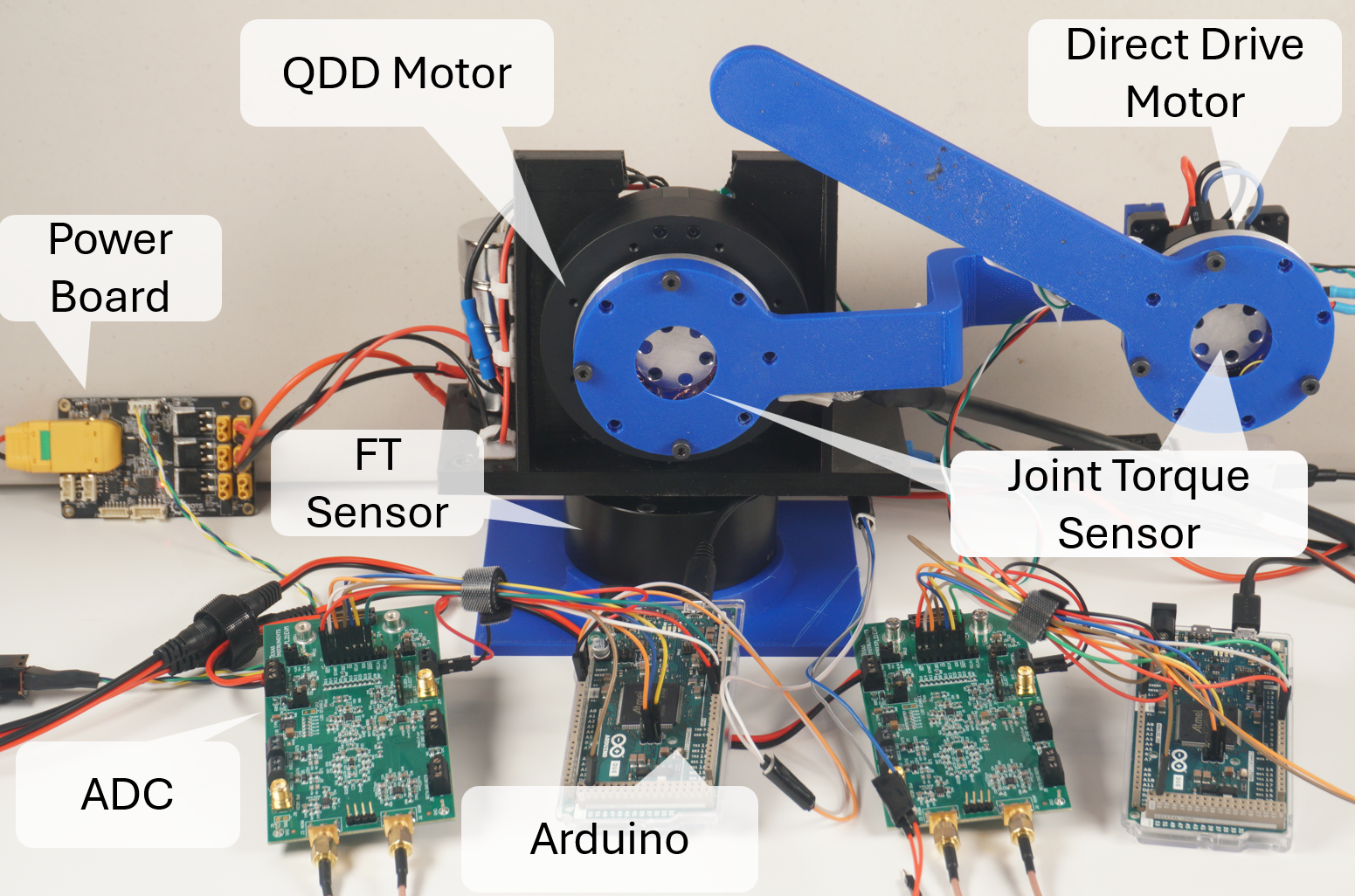}
    \caption{Experimental 2-DoF planar leg testbed with integrated joint torque sensors, data acquisition boards, a hip-mounted Bota Systems PixONE FT sensor, and power distribution board.}
    \label{fig:robot-setup}
    \vspace{-10pt}
\end{figure}

Hardware experiments were performed on a fixed-based 2-DoF planar manipulator with joint torque sensors and mounted on a FT sensor, as shown in Fig.~\ref{fig:robot-setup}, which corresponds to the hardware section of Fig. \ref{fig:flow_diagram} (a). The hip joint is equipped with a quasi-direct-drive (QDD) of mjbots qdd100 (6:1 gearbox) and the knee joint an mj5208 direct drive motor. A torque sensor is mounted on the output shaft of each motor, and the analog outputs from the torque sensors are routed to a 24-bit ADC module (ADS127L21EVM-PDK) via an Arduino~Due communicating over SPI. The ADC transmits data at 2\,MHz baud rate, yielding a sampling rate of 3--4\,kSps. The \textit{mjbots} controllers supply encoder positions $q$ and velocities $\dot{q}$ for each joint while the hip FT sensor data arrives over a separate channel. The data streams are synchronized and forwarded to MATLAB in real time, where the momentum-based collision estimation runs continuously. 

\subsection{Joint Torque Sensor Characterization}
\label{jts_vs_mc}

\begin{figure}[t]
\vspace{5pt}
    \centering
    \includegraphics[width=0.7\columnwidth]{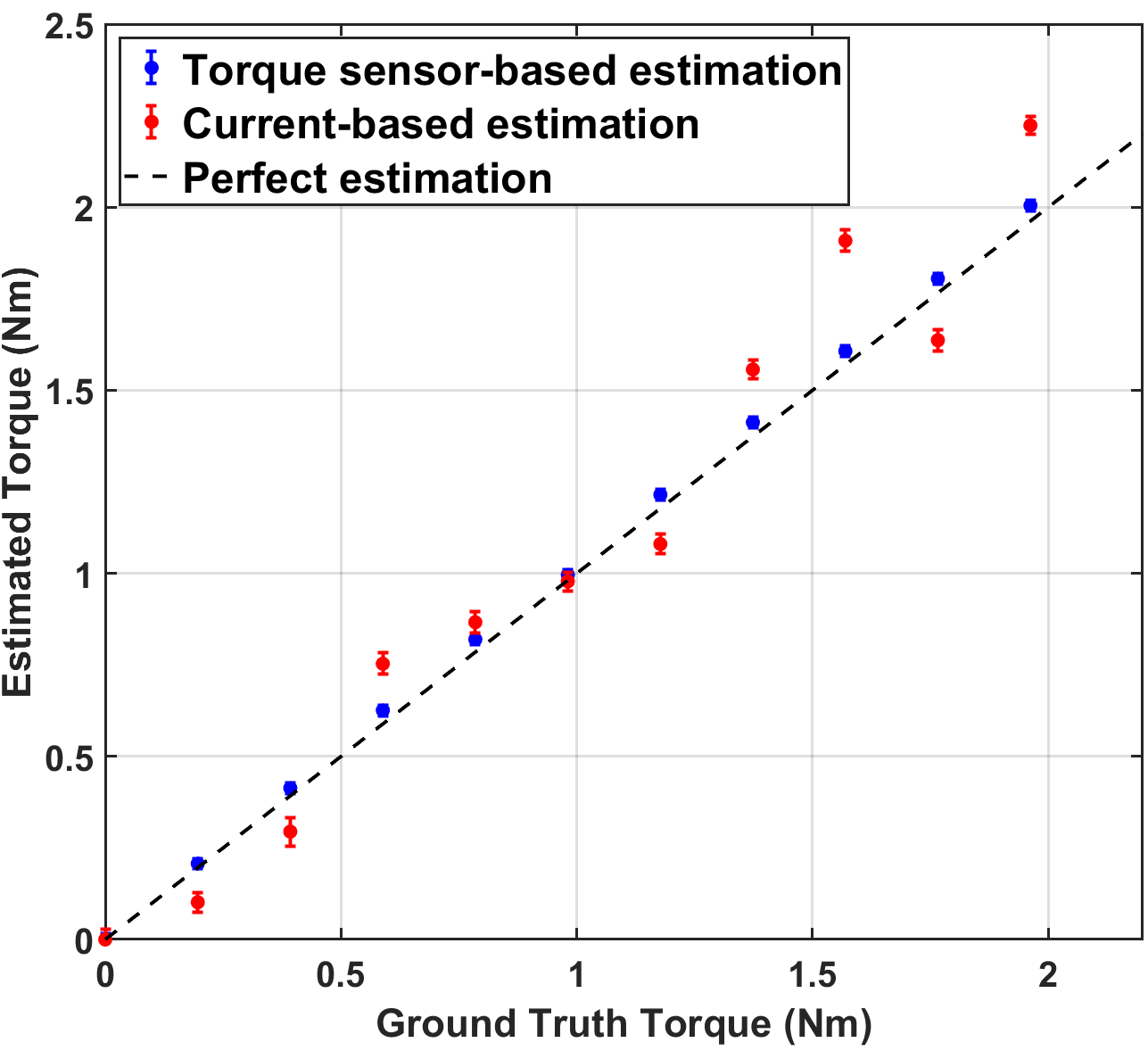}
    \caption{Comparison of joint torque sensor (blue) versus motor current estimation (red) against ground truth torque. The joint torque sensor provides higher accuracy (R$^2$ = 0.9998, RMSE = 0.0317 Nm) compared to motor current estimates (R$^2$ = 0.9609, RMSE = 0.1638 Nm).}
    \label{fig:jts_vs_motor_current_comparison}
    \vspace{-10pt}
\end{figure}

To independently evaluate the custom joint torque sensors, we conducted a standalone calibration and characterization study. A QDD motor was equipped with our joint torque sensor and a lever arm. Known weights from 0.1\,kg to 1\,kg were applied. Direct joint torque sensor measurements and motor-current estimates are plotted against the  ground truth torque at the output shaft and the result is shown in Fig.~\ref{fig:jts_vs_motor_current_comparison}. Both methods perform adequately for moderate loads (0--2\,Nm), while direct torque sensing yields higher linear correlation and lower root-mean-square error (RMSE). In particular, we observe an RMSE of around 0.0317\,Nm and 0.1638\,Nm for the direct joint torque sensor and motor-current approaches, respectively. The joint torque sensors achieve 96.4\% accuracy relative to ground truth torque data. Additional specifications include a mean absolute error (MAE) of 0.0286\,Nm, which represents the practical measurement resolution achievable under typical operating conditions. The sensor design provides a maximum torque capacity of 8.5\,Nm before yielding, as determined through finite element analysis (FEA).

\subsection{Static Loading Tests and Accuracy Evaluation}
To quantify force and location estimation, we conduct static loading by placing known masses at specific positions along each link. Specifically:
\begin{itemize}
    \item \textbf{Link 1 (thigh):} loads of 0.1\,kg and 0.5\,kg, placed at 25\%, 50\%, 75\%, or 100\% of the link length.
    \item \textbf{Link 2 (shank):} loads of 0.05\,kg and 0.1\,kg, similarly positioned at four increments along the link.
\end{itemize}
Due to Link 2's lower torque actuator, smaller test loads were selected. For each placement, the estimated values $\alpha_\mathrm{est}$ and $\mathbf{F}_\mathrm{est}$ are compared against ground-truth values. Table~\ref{tab:hardwareErrors} summarizes the error metrics from the static loading tests. The hardware experiments demonstrate sub-centimeter location accuracy, with Link 2 showing better performance  than Link 1. These values correspond to approximately 4--4.5\% and 2--2.5\% of their respective link lengths. Force estimation errors show expected correlation with load magnitude, maintaining accuracy within a few percent of the applied loads across all test conditions


\vspace{5pt}
\begin{table}[htb]
\vspace{5pt}
    \centering
    \caption{Results of Hardware Experiments}
    \label{tab:hardwareErrors}
    \begin{tabular}{l|c|c|c}
        \hline
        \multirow{2}{*}{\textbf{Configuration}} & \textbf{Load} & \textbf{Location Error} & \textbf{Force Error} \\
        & \textbf{(kg)} & \textbf{RMS (mm)} & \textbf{RMS (N)} \\ 
        \hline
        \multirow{2}{*}{Link 1} & 0.1 & 8.89 & 0.129 \\
        & 0.5 & 7.91 & 0.174 \\
        \hline
        \multirow{2}{*}{Link 2} & 0.05 & 4.09 & 0.045 \\
        & 0.1 & 4.87 & 0.106 \\
        \hline
    \end{tabular}
\end{table}

\section{Discussion and Conclusion}
\label{sec:conclusion}

This paper presents a collision sensing framework that enables legged robots to detect collision anywhere along its links. The collision detection algorithm integrates direct joint torque sensing, a single base force-torque sensor, and a momentum-based observer. This approach allows real-time multi-link collision detection and localization while avoiding complex friction modeling and relying on minimal additional sensing. The proposed approach demonstrates three key contributions validated through experiments: a) a high-bandwidth low-cost joint torque sensor that achieves 96.4\% accuracy relative to ground truth torque at the output shaft, bypassing the need to model complex drivetrain friction; b) a collision detection algorithm using generalized momentum-based observer that fuses joint torque sensor and a base FT sensor; c) hardware experiment of the proposed method verifies that the force and location errors remain within a few millimeters and below 0.2\,N anywhere along the leg. These results demonstrate the potential for safe, responsive operation of dynamic legged robots in cluttered environments.

\section*{Acknowledgments}
We would like to thank Yulun Zhuang, Yue Qin and Yichen Wang for their helpful discussions and feedback on figure design. This project is partially funded by National Science Foundation Graduate Research Fellowship
(DGE-2241144), and National Science Foundation, under grant 2427036.

\vspace{-10pt}

\bibliographystyle{IEEEtran}
\bibliography{reference}

\end{document}